\documentclass[letterpaper]{article} 
\usepackage{aaai2026}  
\nocopyright
\usepackage{times}  
\usepackage{helvet}  
\usepackage{courier}  
\usepackage[hyphens]{url}  
\usepackage{graphicx} 
\usepackage{natbib}  
\usepackage{caption} 
\usepackage{booktabs}
\usepackage[page,title]{appendix} 
\usepackage{adjustbox,makecell,booktabs}

\usepackage{subcaption} 
\usepackage{amsfonts}       
\usepackage{amsmath}
\usepackage{makecell}
\usepackage{amsthm}

\usepackage{comment}
\usepackage{url}
\usepackage{subcaption}
\usepackage{xcolor}

\usepackage{algorithm}
\usepackage[noend]{algpseudocode}
\algnewcommand{\LeftComment}[1]{\Statex \(\triangleright\) #1}
\usepackage{listings}
\DeclareCaptionStyle{ruled}{labelfont=normalfont,labelsep=colon,strut=off} 
\floatstyle{ruled}
\newfloat{listing}{tb}{lst}{}
\floatname{listing}{Listing}
\title{DAO-AI: Evaluating Collective Decision-Making through Agentic AI in Decentralized Governance}
\author {
    Agostino Capponi\textsuperscript{\rm 1},        
    Alfio Gliozzo \textsuperscript{\rm 2},
    Chunghyun Han \textsuperscript{\rm 3},
    Junkyu Lee \textsuperscript{\rm 2}
}
\affiliations{
    \textsuperscript{\rm 1}
    Department of Industrial Engineering and Operations Research and Columbia Business School,\\ Columbia University, New York, NY\\
    \textsuperscript{\rm 2}
    IBM T.J. Watson Research Center, Yorktown Heights, NY\\
    \textsuperscript{\rm 3}
    Department of Industrial Engineering and Operations Research,\\
    Columbia University, New York, NY\\
    ac3827@columbia.edu, gliozzo@us.ibm.com, ch4005@columbia.edu, junkyu.lee@ibm.com
}

\begin{document}
\maketitle
\begin{abstract}
This paper presents a first empirical study of agentic AI as autonomous decision-makers in decentralized governance. Using more than 3K proposals from major protocols, 
we build an \textit{agentic AI voter} 
that interprets proposal contexts, 
retrieves historical deliberation data, 
and independently determines its voting position. 
The agent operates within a realistic financial simulation environment grounded in verifiable blockchain data, implemented through a modular composable program (MCP) workflow 
that defines data flow and tool usage via \texttt{Agentics} framework.
We evaluate how closely the agent's decisions align with the human and token-weighted outcomes, 
uncovering strong alignments measured by carefully designed evaluation metrics. 
Our findings demonstrate that agentic AI can augment collective decision-making by producing interpretable, auditable, and empirically grounded signals in realistic DAO governance settings. 
The study contributes to the design of explainable and economically 
rigorous AI agents for decentralized financial systems.
\end{abstract}

\section{Introduction}
Financial decision-making has traditionally been the domain of centralized institutions such as banks, regulators, and corporate boards who control lending, trading, and asset management. 
These entities operate behind closed doors, often with limited transparency and public accountability. 
In contrast, Decentralized Finance (DeFi) has introduced a radically different model, where financial protocols are governed by global communities through transparent, and programmable systems.

Decentralized Autonomous Organization (DAO), a blockchain-based governance structure that enables collective decision-making without centralized leadership \citep{fritsch2024analyzing,han2025review} is the core of such transformations. 
A DAO is a new organizational form in the digital era, designed to facilitate transparent and autonomous governance through smart contracts. These contracts encode the rules and voting mechanisms of the organization, allowing members to coordinate and allocate resources via on-chain proposal submissions and token-weighted voting.  DAOs provide an alternative to traditional institutions and enables open and verifiable participations in global financial governance by removing intermediaries and enforcing outcomes automatically through smart contracts.

Currently, DAOs govern some of the most influential DeFi platforms in the ecosystem.
For example, \textit{Aave} handles lending and borrowing,  \textit{Uniswap} and \textit{Balancer} manage decentralized exchange and advanced liquidity provisioning.
\textit{Lido} enables liquid staking to keep staked capital usable. and this activity is made efficient by Layer-2 scaling solutions like \textit{Arbitrum} and \textit{Metis}, with \textit{1Inch} aggregating the best trade prices, and \textit{Aura} providing crucial incentives to maximize returns for liquidity providers.

These organizations operate under a shared governance model defined by four core characteristics: (1) decentralized decision-making, where the voting power is distributed among token holders ,
(2) autonomous execution, with the outcomes enforced by smart contracts, (3) transparency and immutability, as all actions are recorded on-chain, and (4) global participation, allowing open and verifiable participations.
These characteristics enable DAOs to collectively manage billions of dollars in treasury funds, with governeance decisions ranging from protocol upgrades to grant allocations having real financial consequences for users, developers, and investors. 

While advances in data and algorithmic governance enhance transparency and efficiency, they also introduce new information asymmetries and participation frictions, as observed in traditional corporate governance research \citep{jiang2024corporate}.
Similar dynamics manifest in decentralized autonomous organizations (DAOs), which face several governance challenges, including low voter participation that frequently falls well below $10\%$ of eligible members, undermining the legitimacy of collective decision-making \citep{laturnus2023economics}.
Such limited engagement often coincides with concentrated voting power, where a small number of large token holders dominate outcomes \citep{AppelGrennan2024}.
Furthermore, governance power is often highly concentrated among a few large token holders,  contradicting the core ethos of decentralization.
This situation gets worse due to the information overload because
the proposals could span complex technical and economic dimensions, making it difficult for the broader community to assess the quality of the proposals, manage cognitive burden, and accurately gauge community sentiment \citep{liu2023illusion,fritsch2024analyzing,chohan2024decentralized,weidener2025delegated}. 
Recent studies also show that governance in decentralized organizations may exhibit persistent power concentration and agenda control \citep{cong_centralized_2025,fan_wisdom_2025,han2023daogovernance, appelControl}. These issues mirror broader problems in financial decision-making. Namely, how to ensure informed, inclusive, and rational outcomes in complex, high-stakes environments. These issues mirror broader problems in financial decision-making including how to ensure informed, inclusive, and rational outcomes in complex, high-stakes environments. 

This paper presents \texttt{DAO-AI}, an agentic AI framework for learning and evaluating decision policies in DAOs. 
Built upon IBM \texttt{Agentics}\footnote{\url{https://github.com/IBM/Agentics}}, \texttt{DAO-AI} orchestrates multiple agents to analyze the metadata of proposals, forum discussions, and voting dynamics. 
Rather than replicating votes, \texttt{DAO-AI} offers an LLM-based decision maker that ingests the proposal context to generate options given the proposals. We empirically evaluate its alignment and generalization across 3,383 proposals from eight major DAOs. 
The framework provides a data-driven decision-making system that enables the simulation of autonomous decisions in open digital economies.

\begin{figure*}
    \centering
    \includegraphics[width=0.99\linewidth]{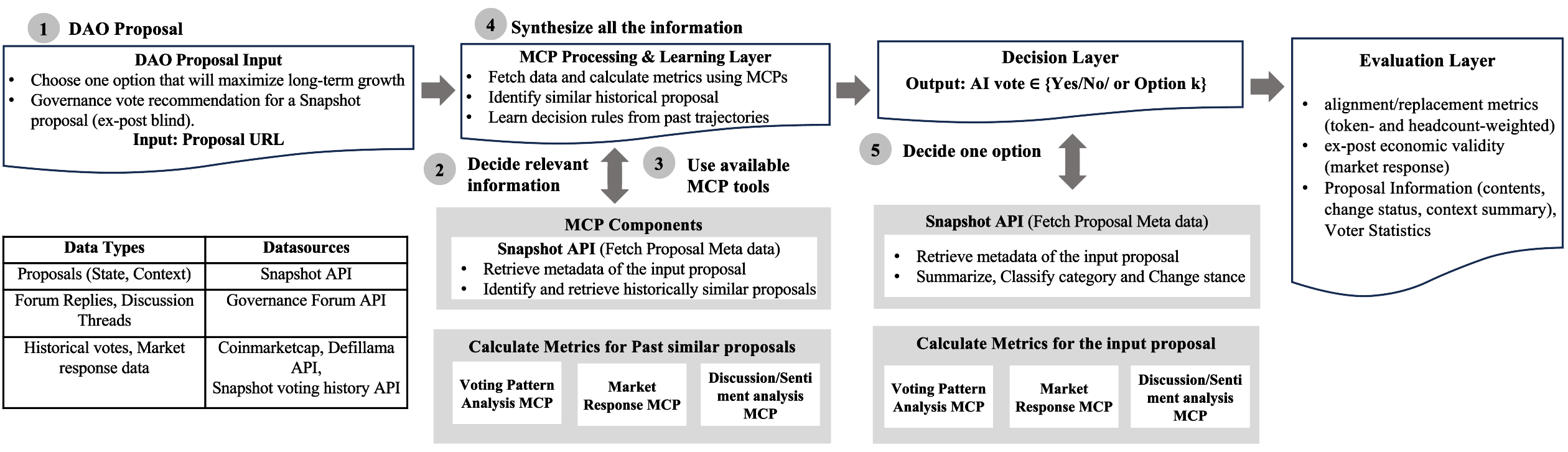}
    \caption{
\footnotesize
\texttt{DAO-AI} Overview:
End-to-end architecture of \texttt{DAO AI} for governance vote recommendation. Given a Snapshot proposal URL as input (1), 
the system orchestrates a sequence of Modular Composable Programs (MCPs) to fetch proposal metadata (2), 
analyze forum discussions, assess market responses, and evaluate voting dynamics (3). The collected evidence is synthesized into a structured prompt, guiding the language model to select a single vote option and generate a justification that reflects community sentiment, historical outcomes, and economic impact (4).     
Finally the system generates a vote recommendation with justification (5).
}\label{fig:overview}
\end{figure*}

\section{Background}
\paragraph{Non-Agentic Approaches to DAO Governance}

Governance in DAOs has historically been driven by human participation supported by static, non-reasoning infrastructure.
The primary goal of these systems is to facilitate token-weighted voting and to manage communal deliberation.
Most DAOs utilize platforms such as \textit{Snapshot.org} or \textit{Tally} to handle proposal interfaces, token-weighted ballot counting, and result finalization.
While these tools successfully enable decentralized participation, they inherently lack automation, structured reasoning, and mechanisms for scalable, consistent proposal evaluation.

The earlier works on improving DAOs have focused on improving the static infrastructure, which have common limitations due to the reliance on human interpretations and involvement.
\begin{itemize}
\item \textbf{Voting Infrastructure:} 
Implement platforms and interfaces for proposal creation and voting execution, without assistance in assessing proposal quality or its potential impact.
\item \textbf{Governance Frameworks:} 
Propose new voting mechanisms aimed at increasing voter participation and ensuring fairness.
\item \textbf{Community Analysis:} 
Conduct human-in-the-loop analysis of the forum discourse or the voter behaviors to understand community sentiment and patterns of influence.
\item \textbf{Organizational Models:} 
Compare DAOs to traditional corporate structures and analyze the inherent agency costs or the governance architecture.
\end{itemize}
Note that the previous approaches do not incorporate autonomous reasoning capabilities, are restricted in their scalability, and offer limited predictive or prescriptive insights into governance outcomes.

\paragraph{Agentic AI in Financial Systems}

The emergence of the agentic AI \footnote{Autonomous systems designed to plan, reason, and execute actions \citep{ng2024agentic}.}
introduces a new paradigm for addressing the aforementioned challenges in DAO governance.  
The agentic AI systems could provide the automated behavior that performs sophisticated tasks essential for complex decision-making, such as perception and interpretation of information by utilizing natural language understanding capability to interpret the technical, economic, and social implications of intricate proposals,
and the language-based reasoning over diverse text data sources, including forum discussions, market data, and on-chain transactional activity. It can also simulate the voting behaviors based on learned preferences in a large amount of data.

\paragraph{AI and Alignment to Human Values}

The primary concern in the deployment of the autonomous systems is \textit{alignment}, which refers to the degree to which an AI system's decisions align with desired human values, preferences, or collective goals.
Aligning with human values becomes particularly crucial in high-stakes environments, such as financial domains.
In the context of DAO governance, the requirement for alignment is unique because the autonomous system in DAO is not tied to the preferences or values of a single user, but rather to complex and collective outcomes involving diverse and often competing stakeholders.

To systematically assess the relationship between the decisions made by the agentic AI system and the human voters,
we adopt a multi-scale alignment framework, as proposed by \citet{capponi2024maximal}. This framework introduces three complementary perspectives for evaluating the system's performance and representativeness, 
ranging from the token-level, voter-level, and the proposal-level. 
We will elaborate the evaluation metrics for the alignment in the following sections. 

\subsection{Agentics Framework}
\texttt{Agentics} is a functional agentic AI framework 
designed for building structured data workflows with LLMs \cite{gliozzo2025transduction}. 
It introduces a data-centric paradigm where agents are defined by typed schemas
called \textit{ATypes} and operate through \textit{logical transduction}, 
which is a structured transformations between input and output types through LLMs. 
In \texttt{Agentics},  each agent is a stateless transducer acting over an 
\textit{Agentic Structure},
$$AG := \{ \text{atype}: \Theta, \text{states}: \text{List}[\text{atype}] \},$$
which is a collection of the typed objects or states, sharing the common $\text{atype} \in \Theta$,
where $\Theta$ denotes the universe of all types.
The \texttt{Agentics} programming model enables asynchronous schema-constrained generation of typed states given input states.
The core operator in \texttt{Agentics}  is the logical transduction,
$$y := AG[Y] \ll x,$$
which maps an input state $x$ of type $X$ to an output state $y$ of type $Y$, ensuring each field in $y$ is transduced from $x$ under the semantic type constraints via LLMs.

\texttt{Agentics} also supports scalable workflows via asynchronous map-reduce operators and quotient structures for abstraction, making it particularly suitable for robust and interpretable financial analytics.

\section{Methodology}
This section presents the design and implementation of \texttt{DAO-AI}, 
a new agentic AI system for simulating and evaluating decentralized governance decisions. 
Built on the \texttt{Agentics} framework \citep{gliozzo2025transduction}, 
\texttt{DAO-AI} leverages the tools to ingest, interpret, and reason over various governance data through Modular Composable Programs (MCPs) protocols \cite{AnthropicMCP}.
The system is designed to answer the following question, 
``Can autonomous agents replicate or improve the quality of collective decision-making in decentralized financial governance?''

\subsection{Agentic Architecture Overview}
\paragraph{DAO General Workflow}
DAO governance generally follows a structured process that begins with (1) Proposal Submission from an eligible member (like a token-holder or delegate). This is followed by a period of (2) Discussion and Review, where the community debates the proposal in public governance forums. The (3) Voting Phase then commences, where token-holders cast their votes, often using off-chain platforms such as Snapshot. For a decision to pass, it must meet predefined (4) Quorum and Thresholds for both participation and approval. Once approved, (5) On-chain Execution occurs, where smart contracts automatically implement the decision, although some DAOs include a (6) Grace or Objection Period allowing members to appeal the decision or exit before it's finalized.

\paragraph{\texttt{DAO-AI} Overview}
\texttt{DAO-AI} simulates the DAO workflow using a modular agentic architecture built on the Agentics framework. 
The system receives a Snapshot proposal URL as input and returns a vote recommendation with justification. 
The architecture is illustrated in Figure~\ref{fig:overview}.
The operational data flow proceeds through the following steps:

\begin{itemize}
    \item \textbf{Input:} 
    The main message prompt provides the input to the system with 
    a Snapshot proposal URL and the instructions to initiate data preparation.

    \item \textbf{Data Preparation:} 
    The data preparation stage determines the relevant data and invokes the appropriate MCP tools.
    \item \textbf{Decision Making:} 
The MCP processing and learning layer constructs a structured prompt using the data collected and synthesized by the data preparation modules. 

In the decision layer, LLM selects a single option that maximizes the long-term growth of the organization. 
The decision prompt also includes several key elements, such as 
the impact of the decision, voting dynamics, 
historical proposal outcomes, and forum sentiment. 
This information signals are integrated into a coherent justification, 
enabling the agent to make a governance recommendation based on natural language reasoning.

    \item \textbf{Output:} 
The final output of the decision-making process includes 
the selected vote option chosen by the agent and the justification of the selection.
\end{itemize}

\subsection{Data Models and MCP Tools}
A core challenge in building LLM-based governance systems is providing 
the ability to access, transform, and reason over various data sources. 
\texttt{DAO-AI} addresses this by leveraging a suite of Modular Composable Program (MCP) tools and logical transduction mechanisms provided by the \texttt{Agentics} framework. 
These tools form the learning layer of the system, enabling data-driven decision-making by transforming raw governance signals into structured abstractions of data.

We employ four MCP tools, each of which is responsible for distinct features that support decision-making, ranging from the collection of factual and historical data to the impact analysis of the decision.

\begin{enumerate}
    \item \textbf{Snapshot MCP (Governance Metadata Agent):} 
    This tool collects the factual and historical data of governance activity from \texttt{Snapshot.org}. 
    It produces two data types, \texttt{ProposalAType}, which captures the metadata in the proposals, such as title, body, options, and timestamps, and \texttt{VoteRecordAType}, which records individual votes and token-weighted votes. We also categorize the proposals into five domains, \textit{Tokenomics}, \textit{Finance}, \textit{Governance}, \textit{Viability}, and \textit{Management} using an LLM-based classification pipeline adapted from \citet{AppelGrennan2024}.

    \item \textbf{Governance Forum MCP (Deliberation Context Agent):} 
    This tool extracts the semantic and reasoning context from the governance forums by processing the discussion thread and performing sentiment analysis. 
The results are stored in \texttt{ForumThreadAType} type that includes sentiment polarity, stance scores, and semantic embeddings.

    \item \textbf{Voting Dynamics MCP (Temporal Participation Agent):} 
    This tool analyzes the temporal change of voting activity using the Snapshot voting history. 
    It computes dynamic features such as lead ratios, spike indices, and participation asymmetries, which are captured in \texttt{ParticipationSeriesAType}. These features help assess the AI alignments.

    \item \textbf{Market Response MCP (Economic Impact Agent):} 
    This tool evaluates the market response and impact of the governance decisions.  
It collects token price and total value locked (TVL) data from CoinMarketCap and DeFiLlama within a $\pm$3-day event window around each proposal. 
    The data is stored in \texttt{MarketAType} type that captures 
the abnormal return and liquidity shifts, enabling the system to link governance outcomes with market sentiment.
\end{enumerate}

Each MCP tool is implemented as a stateless agent that fetches and transforms raw data into the data types, referred to as \textit{ATypes} in \texttt{Agentics}.
Each AType corresponds to a specific aspect of the DAO governance process\footnote{We provide the full description of the data attributes that \texttt{DAO-AI} uses to make decisions in the Appendix.}:
\begin{enumerate}
    \item \textbf{\texttt{ProposalAType}:} 
Captures the metadata from the proposals, such as the unique identifier,  the title, body text, available voting choices, and timestamps for creation and voting windows.
    \item \textbf{\texttt{VoteRecordAType}:} 
Represents the voting records, such as the voter's address, their selected choice(s), and the amount of voting power (typically token-weighted) they exercised.
    \item \textbf{\texttt{ForumThreadAType}:} 
Encodes the semantic context of the forum discussions, such as the proposal ID, the URL of the forum thread, a stance score indicating support or opposition, and a sentiment score indicating positive or negative tone of the discussion.
    \item \textbf{\texttt{MarketAType}:} 
Captures the short-term economic response to a proposal. It includes abnormal changes in token price and total value locked (TVL) within a $\pm$3-day window around the proposal's voting period, serving as a proxy for market sentiment.
\end{enumerate}

\subsection{Logical Transductions for Decision Making}
The \texttt{DAO-AI} system generates governance decisions through a structured process of logical transduction. Given an input proposal and the main message prompt, the system synthesizes typed data from multiple MCP tools
and composes this information into a decision prompt,
which guides the language model to select a single governance option and provide a rationale grounded in structured evidence
\footnote{We provide the full decision prompt in the Appendix.}.

Let the input be an Agentic structure \texttt{AG[P]}, 
where the proposal type schema \texttt{ProposalAType} 
and the typed inputs are collected from MCP tools are defined as follows.
\begin{lstlisting}[language=Python,breaklines=true,showstringspaces=false,basicstyle=\fontsize{7}{7}\ttfamily]
class ProposalAType(BaseModel):
    proposal_id: str
    space_id: str
    title: str
    body: Optional[str]
    choices: List[str]
    created_at: Optional[str]
    start: Optional[str]
    end: Optional[str]
class VoteRecordAType(BaseModel):
    proposal_id: str
    voter: str
    choice: Dict[str, float] | int | List[int] 
    vp: float
class ForumThreadAType(BaseModel):
    proposal_id: str
    url: str
    stance_score: float
    sentiment: float
class MarketAType(BaseModel):
    proposal_id: str
    window_days: int = 3
    price_abnormal: Optional[float]
    tvl_abnormal: Optional[float] 
class DecisionAType(BaseModel):
    selected_option: str
    justification: str    
\end{lstlisting}
These are composed into a message context \texttt{AG[M]}, which aggregates structured signals from voting, forum, and market MCPs. The final decision is generated via logical transduction to \texttt{DecisionAType},
$$
AG[\text{DecisionAType}] \ll AG[M] \ll AG[\text{ProposalAType}].
$$
This composition ensures that the governance decision is transduced from the proposal and its structured context, enabling reproducible and interpretable agentic reasoning.

\section{Dataset and Evaluation Protocol}

Our dataset consists of 3,383 governance proposals sourced from eight major DAOs: Aave, Uniswap, Lido, Balancer, Arbitrum, 1inch, Metis, and Aura. Each proposal is categorized as either \textit{binary}, involving two-option votes such as \textit{For/Against/Abstain} or 
\textit{multiple-choice}, which include three or more options. 
Binary proposals are predominant across most DAOs, 
while multiple-choice votes appear primarily in large, protocol-upgrade decisions.
Table \ref{tab:dao_counts} shows the distribution of the governance proposals across eight DAOs. 
\begin{table}[h!]
\footnotesize
\centering
\caption{Distribution of Governance Proposals across DAOs}
\label{tab:dao_counts}
\begin{tabular}{lccc}
\toprule
\textbf{DAO} & \textbf{Binary} & \textbf{Multiple-choice} & \textbf{Total} \\
\midrule
Balancer              & 777 & 113 & 890 \\
Aura Finance          & 708 & 117 & 825 \\
Aave DAO              & 641 & 88  & 729 \\
Arbitrum              & 302 & 47  & 349 \\
Lido                  & 175 & 22  & 197 \\
Metis                 & 208 & 0   & 208 \\
Uniswap               & 63  & 44  & 107 \\
1inch                 & 78  & 0   & 78  \\
\midrule
\textbf{Total}        & \textbf{2,944} & \textbf{431} & \textbf{3,383} \\
\bottomrule
\end{tabular}
\end{table}

\paragraph{Notation}
For a given proposal $p$ and voter  $i$, 
let $w_{i,p}$ denote the voting power (VP) of voter $i$ on proposal $p$. We define the total voting power as $T_p = \sum_i w_{i,p}$, 
and let $\texttt{vote}_{i,p}$ 
represent the choice made by voter $i$. 
The final outcome of the vote is denoted by $\texttt{final}_p$, 
while $\texttt{AI}_p$ denotes the decision recommended by the AI. 
Let $N^{\texttt{voters}}_p$ be the total number of voters participating in proposal $p$, 
and use $\mathbf{1}\{\cdot\}$ to denote the indicator function.

\subsection{Evaluation of AI Alignments}
To assess the quality and representativeness of decisions made by \texttt{DAO-AI}, 
we evaluate its alignment with human governance behavior at two distinct levels. 
At the token level, we examine whether the decision by AI aligns with the token-weighted majority outcome or with the behavior of a typical human voter, reflecting how well the AI captures the preferences of economically influential participants. 
At the voter level, we assess whether the decision by AI aligns with the majority of individual voters or with the behavior of a typical participant, providing insight into whether the AI acts as a representative of the community. Next, we introduce formal metrics to quantify alignment at both levels.

\paragraph{Token-level Alignment}
To evaluate token-level alignment, we begin by defining the economic majority share,
\begin{equation}
S_p = \frac{\sum_{i} w_{i,p}\,\mathbf{1}\{\texttt{vote}_{i,p} = \texttt{final}_p\}}{T_p},    
\end{equation}
which measures the fraction of total voting power that supported the winning option on proposal $p$
in the collected dataset.

Next, we define the AI token-weighted alignment,
\begin{equation}
A^{\texttt{AI}}_p = \frac{\sum_{i} w_{i,p}\,\mathbf{1}\{\texttt{vote}_{i,p} = \texttt{AI}_p\}}{T_p},
\end{equation}
which captures the share of total voting power that voted in agreement with the AI's recommendation. 
Averaging this across all proposals in the evaluation set $P$,
we compute the global AI token alignment,
\begin{equation}
\overline{A}^{\texttt{AI}} = \frac{1}{|P|} \sum_{p \in P} A^{\texttt{AI}}_p.
\end{equation}
To benchmark this alignment against human behavior, we define the human token-weighted agreement
for each voter $i$,
\begin{equation}
\tilde{A}_i = \frac{\sum_{p \in P_i} w_{i,p}\,\mathbf{1}\{\texttt{vote}_{i,p} = \texttt{final}_p\}}{\sum_{p \in P_i} w_{i,p}},    
\end{equation}
which measures how often their votes aligned with the final outcome, weighted by their voting power across the proposals they participated in.

Finally, we compare the AI’s global alignment score $\overline{A}^{\texttt{AI}}$ 
to the median of the human-token-weighted scores $\mathrm{median}(\{\tilde{A}_i\})$.
We say \texttt{DAO-AI} decision aligns more closely with the token-weighted majority than a typical human voter
if the following condition holds
\begin{equation}
\overline{A}^{\texttt{AI}} > \mathrm{median}(\{\tilde{A}_i\}).
\end{equation}
This comparison provides a benchmark for evaluating whether 
\texttt{DAO-AI} behaves like a rational delegate aligned with the prevailing economic consensus.

\paragraph{Voter-level Alignment}
To evaluate voter-level alignment, 
we define metrics based on headcount rather than token weight. 
First, we compute the AI headcount alignment for each proposal $p$, 
which measures the fraction of individual voters whose choice matched the AI's decision,
\begin{equation}
H^{\texttt{AI}}_p = \frac{\big|\{\,i:\ \texttt{vote}_{i,p} = \texttt{AI}_p\,\}\big|}{N^{\texttt{voters}}_p}.
\end{equation}

Next, we define the human headcount benchmark for each voter $i$, 
which captures the fraction of proposals where the voter chose the realized winner,
\begin{equation}
\hat{A}_i = \frac{1}{|P_i|} \sum_{p \in P_i} \mathbf{1}\{\texttt{vote}_{i,p} = \texttt{final}_p\}.
\end{equation}

Averaging $H^{\texttt{AI}}_p$ across all proposals 
in the evaluation set yields the global AI headcount alignment $\overline{H}^{\texttt{AI}}$. 
We then compare this value to the median of the human headcount benchmarks $\{\hat{A}_i\}$. 
We say that \texttt{DAO-AI} aligns more closely with the voter majority than a typical human participant if the following condition holds:
\begin{equation}
\overline{H}^{\texttt{AI}} > \mathrm{median}(\{\hat{A}_i\}).
\end{equation}
This comparison provides a benchmark for evaluating whether 
the decision by \texttt{DAO-AI} align with the majority of voters more consistently than the median human participant.

\section{Experiments}

We evaluate the proposed agentic AI on 3,382 DAO proposals. 
Our analysis focuses on two research questions.
\begin{itemize}
    \item \textbf{RQ1}: Does \texttt{DAO-AI} align more consistently with collective DAO outcomes than the human voters? 
    \item \textbf{RQ2}: Does \texttt{DAO-AI} predicts better decisions that 
    subsequently led to the positive outcomes?
\end{itemize}

\paragraph{RQ1: Does \texttt{DAO-AI} align more closely with collective decisions?}

Across all proposals, 
the AI’s simulated decisions coincide with the final DAO outcomes in 92.5\% of cases, 
compared to the average human voter’s agreement rate of 76.6\%. 
This advantage persists under both token- and headcount-weighted definitions of majority, 
with the proposal-level averages of $A^{AI}_{token}=0.91$ and $H^{AI}_{p}=0.91$,
as shown in Table \ref{tab:overall-stats}
The high economic-majority share $S = 0.94$ indicates that many proposals are decisive.
To ensure this result is not driven solely by easy cases, 
we conduct robustness checks on contested proposals and participation-filtered subsets. 
Across these stress tests, the relation 
$\tilde{A}^{\text{AI}} \ge \tilde{A}^{\text{Human}}_{\text{avg}}$ 
remains stable: the median proposal-level alignment of the AI agent 
($\tilde{A}^{AI}_p = 0.99$) exceeds that of human voters 
($\tilde{A}^{Human}_p = 0.96$), and the aggregate mean alignment 
($\bar{A}^{AI} = 0.912$ vs.\ $\bar{A}^{Human} = 0.908$) 
shows consistent superiority across evaluation subsets.

\begin{table}[H]
\centering
\caption{Aggregate Statistics across Proposals}
\label{tab:overall-stats}
\setlength{\tabcolsep}{5pt}
\footnotesize
\begin{adjustbox}{max width=1\linewidth}
\begin{tabular}{@{}lcccccc@{}}
\toprule
\textbf{Metric} & \textbf{Mean} & \textbf{Median} & \textbf{Std} & \textbf{Q25} & \textbf{Q75} & \textbf{Max} \\
\midrule
$A^{AI}_p$  & 0.91 & 0.99 & 0.22 & 0.98 & 1.00 & 1.00 \\
$H^{AI}_p$   & 0.91 & 0.96 & 0.15 & 0.90 & 1.00 & 1.00 \\
$S_p$   & 0.94 & 0.99 & 0.15 & 0.98 & 1.00 & 1.00 \\
$N_{\text{voters}}$ & 1142.84 & 145 & 1904.11 & 10 & 1163 & 5500 \\
\bottomrule
\end{tabular}
\end{adjustbox}
\end{table}

\noindent
\textbf{Interpretation:} The AI’s vote choices mirror collective decisions more reliably than the average human participants, supporting its potential as a “representative” in decentralized voting.

\vspace{-1.5mm} 
\begin{table}[H]
\footnotesize
\centering
\caption{AI vs.\ Humans: Agreement with Final Decision by the Decision Type.}
\label{tab:ai-vs-human-final}
\renewcommand{\arraystretch}{1.12}
\begin{adjustbox}{max width=1\linewidth}
\begin{tabular}{lcccc}
\toprule
\textbf{Bucket} & \textbf{N} & \textbf{Humans} & \textbf{AI} & \textbf{Difference (pp)} \\
\midrule
\makecell[l]{Binary\\(change = yes)}         & 2553 & 0.9427 & 0.9909 & +4.83 \\
\makecell[l]{Binary\\(change = no)}          &  398 & 0.6121 & 0.6373 & +2.52 \\
\makecell[l]{Multi-option\\(change = yes)}   &  360 & 0.8333 & 0.8914 & +5.81 \\
\makecell[l]{Multi-option\\(change = no)}    &   71 & 0.4213 & 0.3571 & -6.42 \\
\bottomrule
\end{tabular}
\end{adjustbox}
\raggedright\footnotesize\emph{Note.} \textit{change = yes} indicates that the proposal content calls for implementing a protocol change
\end{table}

\vspace{-1.3em}
\paragraph{RQ2: Does \texttt{DAO-AI} exhibit ex-post validity in the economic outcomes?}
We assess whether the AI’s decisions align with \textit{ex-post} economic outcomes. 
Specifically, whether the options it supported were followed by favorable market reactions. 
Across all proposals, 
the probability that \texttt{DAO-AI} endorsed decision was followed 
by a positive price or TVL (Total Value Locked) response is 
\begin{itemize}
    \item $P(\Delta P>0|\text{AI})\in [0.46, 0.52]$
    \item $P(\Delta \text{TVL}>0|\text{AI})\in [0.40, 0.55]$,
\end{itemize}
closely matching the baselines observed from all human-adopted proposals, 
\begin{itemize}
    \item $P(\Delta P>0|\text{Final}) \in [0.46, 0.48]$
    \item $P(\Delta \text{TVL}>0|\text{Final})\in [0.40, 0.54]$,
\end{itemize}
as shown in Table \ref{tab:cond-prob}.
These results indicate that the decisions by \texttt{DAO-AI} are not random but economically consistent with outcomes that the market rewards.

\textbf{Interpretation:}
AI-endorsed decisions exhibit comparable or slightly higher ex-post success rates relative to human-adopted baselines, indicating that the agent's voting logic is economically valid rather than random or purely mimetic.

\begin{table}[H]
\centering
\caption{Conditional Probability of the Positive Ex-post Responses by Proposal Type and Decision Direction.}
\label{tab:cond-prob}
\begin{subtable}[t]{\linewidth}
\centering
\caption{Price responses}
\begin{adjustbox}{max width=\linewidth}
\begin{tabular}{lcc}
\toprule
\textbf{Proposal Type} & $P(\Delta P>0|\text{AI})$ & $P(\Delta P>0|\text{Final})$ \\
\midrule
Binary (change = yes)      & 0.459 & 0.457 \\
Binary (change = no)       & 0.522 & 0.479 \\
Multi-option (change = yes)& 0.497 & 0.460 \\
Multi-option (change = no) & 0.320 & 0.471 \\
\bottomrule
\end{tabular}
\end{adjustbox}
\end{subtable}

\vspace{0.5em} 

\begin{subtable}[H]{\linewidth}
\centering
\caption{TVL responses
}
\label{tab:expost_tvl}
\begin{adjustbox}{max width=\linewidth}
\begin{tabular}{lcc}
\toprule
\textbf{Proposal Type} & $P(\Delta TVL>0|\text{AI})$ & $P(\Delta TVL>0|\text{Final})$ \\
\midrule
Binary (change = yes)      & 0.516 & 0.514 \\
Binary (change = no)       & 0.399 & 0.403 \\
Multi-option (change = yes)& 0.553 & 0.540 \\
Multi-option (change = no) & 0.480 & 0.543 \\
\bottomrule
\end{tabular}
\end{adjustbox}
\end{subtable}
\end{table}
\noindent

\vspace{-1.3em}
\section{Robustness and Sensitivity Analyses}
\label{sec:robustness}
To ensure that the observed performance of \texttt{DAO-AI} is not driven by favorable conditions of the dataset, we conduct robustness and sensitivity analyses. 

\paragraph{Contested-only (Close Votes)}
To assess performance in difficult cases, we isolate proposals with weak consensus, defined by an economic majority share \( S_p \le 0.60 \). These “contested” proposals represent situations where token-weighted support is fragmented and the final outcome is less decisive.
Alignment drops as expected, revealing that \texttt{DAO-AI} also performs worse in difficult cases.

\begin{table}[h!]
\footnotesize
\centering
\caption{Contested-only results ($S_p \le 0.60$).}
\label{tab:contested}
\renewcommand{\arraystretch}{1.12}
\begin{tabular}{lcc}
\toprule
\textbf{Metric} & \textbf{Estimate} & \textbf{N} \\
\midrule
$P(\mathrm{AI}=\mathrm{final})$ & $0.542$ & $156$ \\
$\overline{A}^{AI}$ (token)     & $0.453$ & $156$ \\
$\overline{H}^{AI}$ (headcount) & $0.668$ & $156$ \\
$\overline{S}$                  & $0.374$ & $156$ \\

\bottomrule
\end{tabular}
\end{table}

\vspace{-1em}
\paragraph{Stable Human Baselines}
In practice, not all voters participate regularly. 
Such \emph{participation sparsity} can inflate or deflate per–voter agreement rates, thus biasing the aggregate human benchmark and increasing variance. 
To mitigate this, we construct a stable baseline 
by retaining voters with sufficient participation, $|P_i| \ge 5$. 
Under this criterion, the average human agreement with the final decision is $0.766$.

\paragraph{Interpretation of Robustness Results}
Restricting to contested proposals removes easy, near-unanimous cases and provides a conservative lower bound on performance.
In this difficult subset we observe that
$P(\text{AI}=\text{final})=0.542$,
$\overline{A}^{AI}=0.453$ (token),
$\overline{H}^{AI}=0.668$ (headcount), and
$\overline{S}=0.374$ ($N=156$).

First, headcount alignment remains relatively strong ($0.668$) compared to token-weighted alignment ($0.453$), suggesting that when token power is fragmented, 
\texttt{DAO-AI} tends to stay closer to the \emph{voter} majority than to large-token coalitions. 
Second, 
 binary proposals show higher AI–final agreement ($0.581$, $N=118$)
 than multi-option proposals ($0.421$, $N=38$), 
 consistent with the coordination ambiguity in multi-choice settings.

Taken together, these robustness checks confirm that the overall alignment pattern is not solely driven by easy cases, and that \texttt{DAO-AI} retains a comparative advantage over typical human voters even under stricter evaluation conditions.

\paragraph{Temporal Decision Modeling and Ex-ante Simulation}

To assess whether access to deliberation signals materially affects the AI’s voting outcomes, 
we compare the agent’s decisions at two temporal points: the proposal's inception (ex-ante) and the final stage of voting (ex-post). 
In the ex-ante setting, the agent relies solely on static proposal information such as its content, the proposer's reputation, and historical analogs, without access to ongoing discussions or intermediate vote tallies. Across the entire proposal set, 3.97\% (\textit{N} = 134) of proposals exhibited different choices between the first and last voting points. This alignment with actual outcomes remains nearly identical, with 91.68\% for ex-ante and 92.45\% for ex-post decisions. 
This comparison shows that the AI’s voting behavior remains largely consistent across temporal settings, 
demonstrating that its policy generalizes well even in the absence of deliberation signals. 
However, the ex-post decisions exhibit a modest but meaningful improvement in alignment, 
indicating that additional deliberation and updated contextual information help the agent refine its judgments. 
Overall, these results suggest that while \texttt{DAO-AI} makes stable decisions based on latent proposal features, its performance can benefit from the incremental information available later in the voting process.

\begin{table}[H]
\setcellgapes{3pt}
\makegapedcells
\centering
\caption{Aggregate alignment and outcome metrics}
\vspace{2pt}
\footnotesize
\renewcommand{\arraystretch}{1}
\begin{tabular}{lcc}
\hline
\textbf{Metric (mean)} & \textbf{Ex-ante} & \textbf{Ex-post} \\
\hline
$P(\mathrm{AI} = \mathrm{final})$ & 0.9168 & 0.9245 \\
$\overline{A}^{AI}$ (token) & 0.8943 & 0.9120 \\
$\overline{H}^{AI}$ (headcount) & 0.8999 & 0.9083 \\
$\overline{S}$ & 0.9409 & 0.9739 \\
\hline
\end{tabular}
\end{table}

\vspace{-1em}
\paragraph{Scalability and Agentics Framework}
While our main evaluation covers eight protocols, the importance of scalability grows rapidly as the ecosystem expands. Evaluating AI–human alignment across hundreds of protocols, thousands of proposals, and hundreds of millions of individual votes presents a significant data-processing challenge. Manual human analysis or ad-hoc scraping pipelines become infeasible in such settings. 
To address scale-out challenges, 
the \texttt{Agentics} framework offers high-throughput performance through
(1) parallelized MCP agents that retrieve and normalize proposal data across DAOs,
(2) asynchronous batching and caching to enable efficient, large-scale LLM-based classification, and
(3) distributed vote aggregation and outcome normalization across compute nodes.

\section{Conclusion and Future Work}
This paper presents a preliminary exploration into the potential of agentic AI 
to enhance collective decision-making.
Decentralized Autonomous Organizations 
offer a unique experiment setting
because their governance artifacts such as 
proposal texts, vote records, timestamps, and outcomes are all publically accessible.
At the same time, several communities are actively experimenting with AI delegate agents (e.g., proposals for personalized AI voting agents in Arbitrum), highlighting the timeliness \cite{arbitrum_ai_delegation_2024}.

We introduce a data-centric, agentic AI framework in which 
governance evidence is normalized into typed abstractions (ATypes), 
ingested by modular MCP agents, and transformed via logical transductions into AI generated votes and auditable evaluation metrics. 
We aim to provide a transparent baseline for future advances in data, tooling, and model quality, not to claim a complete solution.

Our empirical results show strong alignment between the agent's 
simulated decisions and the realized outcomes, e.g., $P(\text{AI}=\text{final})\approx0.93$).
Under a per-voter replacement test, we show that
the agent outperforms the average human’s agreement rate ($0.93$ vs.\ $0.77$), 
supporting its viability as an average-voter substitute. Beyond alignment, we further demonstrate that AI-endorsed decisions exhibit \textit{ex-post economic validity}. They are followed by market or protocol responses comparable to, or even exceeding, those of human-approved proposals, suggesting that agentic AI can enhance the efficiency of decentralized decision-making.

\paragraph{Future work} We outline three complementary directions to advance both engineering performance and economic rigor
of agentic AI in decentralized governance.

\begin{enumerate}
  \item \textbf{Data and Scope Expansion} 
  Extend beyond the eight protocols to a broader and more diverse set of DAOs. This includes stratifying by protocol category to study domain-specific failure modes,
  increasing temporal coverage, and incorporating richer deliberation corpora. 
  
  \item \textbf{Tooling and Model Improvements} 
  \begin{itemize}
      \item \emph{Better tooling}: richer MCPs for on-chain traces, validator behavior, and market microstructure.
      Introduce retrieval policies conditioned on proposal type
      and temporal participation features for warning signals. 
  \item \emph{Model upgrades}: 
  compare and ensemble multiple LLMs (beyond the current \texttt{gpt-4o-mini} configuration), 
  and explore instruction optimization with self-play and transduction-guided search.
  \item \emph{Evaluation tooling}: 
  standardized contested buckets, adversarial counterfactual prompts, and human-in-the-loop audits of error clusters.
  \end{itemize}
  
  \item \textbf{Economic Rigor and Causal Identification.}
  Move from alignment reporting to policy-relevant impact estimation. We will employ difference-in-differences (DID) designs to estimate the economic benefit of agentic AI adoption, controlling for latent trends and protocol fixed effects. 

\end{enumerate}

\paragraph{Limitations}
The current study is observational and subject to limitations such as
(1) the coverage of publicly available governance data,
and (2) reliance on a single primary LLM configuration.

\bibliography{ref}

\appendix
\onecolumn
\setcounter{secnumdepth}{2}

\section{Proposal Example from Uniswap and Poll}
\begin{figure}[h]
    \centering
    \includegraphics[width=0.99\linewidth]{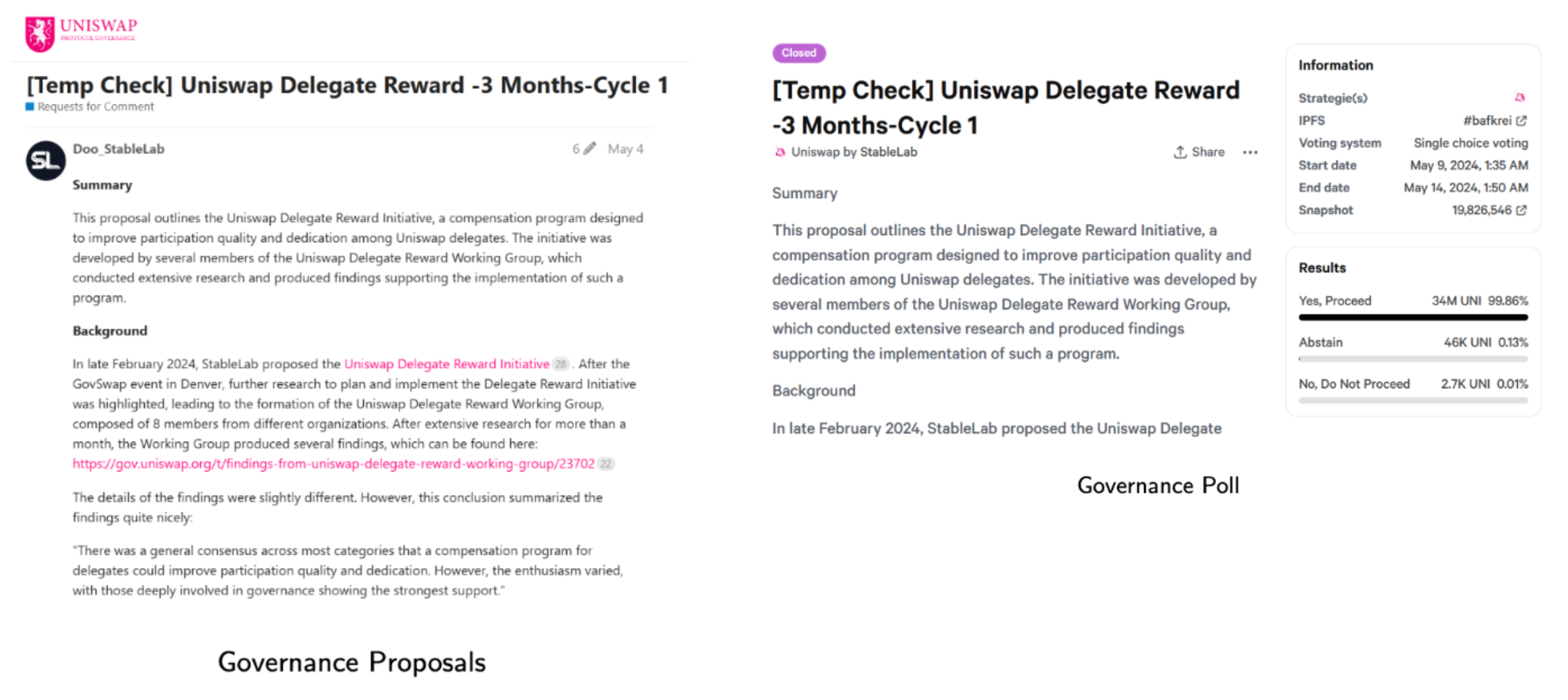}
    \caption{
    Proposal Example from Uniswap and Poll.
}\label{fig:proposal}
\end{figure}

\section{Proposal Categories}
\begin{table}[h!]
\centering
\small
\caption{The categories are derived from LLM-based classification of proposal contents, where the model grouped proposals into ten functional areas of DAO governance.}
\vspace{2pt}
{\renewcommand{\arraystretch}{1.3}
\setlength{\arrayrulewidth}{0.4pt}
\begin{tabular}{|p{0.22\linewidth}|p{0.60\linewidth}|p{0.14\linewidth}|}
\hline
\textbf{Category} & \textbf{Description} & \textbf{N (\%)} \\
\hline
\textbf{Gauge Management} &
Proposals to adjust and manage liquidity parameters and asset support—loan-to-value ratios, liquidation thresholds, and new borrowing assets—to enhance capital efficiency. &
781 (23.1\%) \\ \hline
\textbf{Funding Proposals} &
Initiatives seeking financial resources or support for projects within the DAO, including technology deployment, asset onboarding, or creation of management roles. &
470 (13.9\%) \\ \hline
\textbf{Token Integration} &
Proposals adding new tokens as collateral, whitelisting protocols, or adjusting token functionalities to improve user experience and platform diversity. &
408 (12.1\%) \\ \hline
\textbf{Incentive Program} &
Proposals that create or modify incentive structures, rewarding contributors, improving liquidity, or promoting specific ecosystem initiatives. &
307 (9.1\%) \\ \hline
\textbf{Stablecoin Integration} &
Proposals integrating or enhancing stablecoins in the ecosystem—onboarding new ones, upgrading functionalities, and improving usability and efficiency. &
290 (8.6\%) \\ \hline
\textbf{Risk Management} &
Proposals identifying and mitigating risks through risk oracles, safety module updates, and external partnerships to strengthen DAO security and stability. &
280 (8.3\%) \\ \hline
\textbf{Ecosystem Development} &
Proposals aimed at enhancing and expanding the ecosystem surrounding a decentralized platform. Includes deploying new technologies, establishing partnerships, and improving existing services to foster growth and adoption. &
266 (7.9\%) \\ \hline
\textbf{Treasury Management} &
Proposals managing DAO financial resources—treasury allocation, asset listing, and governance frameworks—to optimize capital utilization and operational efficiency. &
251 (7.4\%) \\ \hline
\textbf{Liquidity Management} &
Proposals optimizing liquidity within DeFi protocols, funding liquidity committees, adjusting reserves, and introducing new liquidity mechanisms. &
210 (6.2\%) \\ \hline
\textbf{Ecosystem Integration} &
Proposals integrating new assets and services into a DeFi ecosystem, onboarding tokens as collateral, enhancing liquidity, and expanding protocol utility. &
120 (3.6\%) \\ \hline
\end{tabular}
}
\end{table}

\newpage

\section{Data Types}
The following Listing shows the full list of the data attributes that \texttt{DAO-AI} uses to make decisions.
\begin{lstlisting}[language=Python,breaklines=true,showstringspaces=false,basicstyle=\fontsize{7}{7}\ttfamily,caption=AType Declarations in Pydantic Model]
class ProposalAType(BaseModel):
    proposal_id: str
    space_id: str
    title: str
    body: Optional[str]
    choices: List[str]
    created_at: Optional[str]
    start: Optional[str]
    end: Optional[str]

class VoteRecordAType(BaseModel):
    proposal_id: str
    voter: str
    choice: Dict[str, float] | int | List[int] 
    vp: float                                 

class ForumThreadAType(BaseModel):
    proposal_id: str
    url: str
    stance_score: float        # pro vs. con ([-1,1])
    sentiment: float           # polarity ([-1,1])

class MarketAType(BaseModel):
    proposal_id: str
    window_days: int = 3
    price_abnormal: Optional[float]  # vs. market index
    tvl_abnormal: Optional[float]

class SnapshotMCP:
    def fetch_proposals(self, space_ids: List[str]) -> List[ProposalAType]: ...
    def fetch_votes(self, proposal_id: str) -> List[VoteRecordAType]: ...

class ForumMCP:
    def fetch_threads(self, proposal_id: str) -> List[ForumThreadAType]: ...

class TimelineMCP:
    def build_series(self, votes: List[VoteRecordAType]) -> ParticipationSeriesAType: ...

class MarketMCP:
    def fetch_market(self, proposal_id: str) -> MarketAType: ...
\end{lstlisting}

\section{Prompts}
\subsection{Main Message Prompt}
The instruction template with the main input message is shown below. 
\begin{lstlisting}[language=,breaklines=true,showstringspaces=false,basicstyle=\fontsize{7}{7}\ttfamily,caption=Main Message Prompt with Proposal]
## Main Message: Governance vote recommendation for a Snapshot proposal (ex-post blind).
- Choose exactly one option from the available choices. 
- If your choice diverges from the leading tally, assume your guidance prioritizes the organization's durable benefit.
- Use available MCP tools to gather forum discussions and run sentiment analysis to report total comments.
- Find similar historical proposals, summarize their impact, and treat declines as unsuccessful lessons to inform the future.
- Clarify whether vote does or does not reflect community views, and whether the recommendation mirrors aggregated input.
- Synthesize vote counts, analytics, and sentiment to explain.
\end{lstlisting}

\subsection{Decision Prompt}
The instruction template used for decision-making is shown below. 
\begin{lstlisting}[language=,breaklines=true,showstringspaces=false,basicstyle=\fontsize{7}{7}\ttfamily,caption=Decision Prompt]
## Objective: Choose exactly one option from the proposal's choices that will maximize the organization's long-term growth
* Impact Reminder: (Proposal, Voting MCPs)
- Assume your recommendation could change the final tally.
- Use vote progress only as a data point, but make an independent choice that maximizes the organization's long-term growth.
* Voting Pattern: (Timeline MCPs)
- Evaluate whether the temporal voting pattern indicates stable consensus formation or late-stage volatility in the decision process.
* Historical Lessons: (Snapshot Proposal MCPs), (Coinmarketcap, Defillama MCPs)
- Retrieve similar past proposals.
- Note whether post-vote token price or TVL declined.
- Treat declines as unsuccessful outcomes and extract lessons.
* Sentiment Alignment: (Sentiment Analysis MCPs)
- Inspect forum discussion comments posted before proposal end only.
- Judge whether aggregated forum sentiment supports or opposes the likely vote outcome.
- Count of positive, negative, neutral sentiments of Comments
* Integration:
- Weave lessons from similar proposals and forum sentiment counts into ai_final_reason, alongside market and timeline analytics.
\end{lstlisting}

\section{Voting Dynamics}

The Voting Dynamics MCP derives proposal level temporal features summarizing collective voting dynamics. 
Metrics are emitted with the \texttt{ParticipationSeriesAType}.

\begin{itemize}
  \item \textbf{Quartile lead ratios} measures how often option $i$ led within quartile $q$ of the voting window.
  \begin{equation*}
  \text{lead\_ratio\_by\_quartile}_{q,i} = 
  \frac{\text{leadHits}_{q,i}}{\max(1,\ \text{votes}_{q})}
  \end{equation*}
  
  \item \textbf{Lead ratio}  
  \begin{equation*}
  \text{lead\_ratio\_total}_i =
  \frac{\text{leadHitsTotal}_i}{\sum_j \text{leadHitsTotal}_j}
  \end{equation*}

  \item \textbf{Early support} isolates first–quartile leadership strength, indicating how strongly option $i$ dominated the early stage of voting.
  \begin{equation*}
  \text{early\_ratio}_i =
  \frac{\text{leadHits}_{\text{Q1}, i}}
  {\max(1,\ \sum_j \text{leadHits}_{\text{Q1}, j})}
  \end{equation*}

  \item \textbf{Spike and stair metrics} quantify short surges and sustained accumulation patterns.
  \begin{equation*}
  \text{spike\_index} =
  \frac{\max(\text{votePowerStep})}{\text{winnerTotalVP}}
  \end{equation*}
  \begin{equation*}
  \text{spike\_follow\_support\_ratio} =
  \frac{\text{winner-aligned VP after spike}}
  {\text{total VP after spike}}
  \end{equation*}
  \begin{equation*}
  \text{stairwise\_ratio} =
  1 - \frac{\text{top decile VP mass}}{\text{winnerTotalVP}}
  \end{equation*}

  \item \textbf{Half-slope difference} compares late vs.\ early accumulation slopes. 
  \begin{equation*}
  \text{half\_slope\_diff} =
  \overline{\Delta \text{VP}}_{\text{late}} -
  \overline{\Delta \text{VP}}_{\text{early}}
  \end{equation*}

  \item \textbf{Meta counts.}  
  Unique voters, total votes, first/last timestamps, and per–quartile voting–power sums accompany each metric bundle.

\end{itemize}

The \textbf{Market Response MCP} evaluates how market indicators react to governance decisions using an event window of $\pm3$ days around each proposal outcome. 
It integrates data from the DeFiLlama API (TVL and treasury values) and CoinMarketCap API (token prices), computing windowed and abnormal percentage changes for each metric. 
All outputs are stored in \texttt{MarketAType} with standardized timestamps and protocol identifiers.

\begin{itemize}

    \item \textbf{Token price impact.}  
    Market reactions are measured as windowed percentage change:
    \begin{equation*}
        \Delta\%_{\text{price}} = 
        \Biggl(\frac{\text{postAvg}}{\text{preAvg}} - 1\Biggr) \times 100
    \end{equation*}
    where \textit{preAvg} and \textit{postAvg} denote mean token prices in pre- and post-event segments. 
    Optionally, an aggregate crypto market index (e.g., CMC Top~100) is subtracted to compute abnormal performance.

     \item \textbf{Market-adjusted abnormal return.}  
    To control for broad market movements, DAO~AI computes an adjusted excess return using the S\&P~Digital~Assets~Index:
    \begin{equation*}
        \text{adjReturn} =
        \Biggl[\Bigl(\frac{\text{postAvg}_{\text{token}}}{\text{preAvg}_{\text{token}}} - 1\Bigr)
        - \Bigl(\frac{\text{postAvg}_{\text{S\&PDA}}}{\text{preAvg}_{\text{S\&PDA}}} - 1\Bigr)\Biggr] \times 100
    \end{equation*}
    isolating token-specific market responses from overall digital-asset market trends.

    \item \textbf{DeFiLlama TVL change.}  
    The MCP compares pre- and post-event total value locked (TVL) to quantify liquidity responses:
    \begin{equation*}
        \text{abnormal\_TVL\_change} =
        \frac{\text{postAvg}_{\text{TVL}}}{\text{preAvg}_{\text{TVL}}} - 1
    \end{equation*}
    representing percentage growth or contraction in TVL relative to the pre-event baseline.

    \item \textbf{Treasury variation.}  
    Treasury responses are computed analogously:
    \begin{equation*}
        \text{abnormal\_treasury\_change} =
        \frac{\text{postAvg}_{\text{treasury}}}{\text{preAvg}_{\text{treasury}}} - 1
    \end{equation*}
    capturing shifts in on-chain reserves or protocol-controlled assets around the governance event.

\end{itemize}

\end{document}